\pdfoutput=1

\documentclass[11pt]{article}

\usepackage[final]{acl}

\usepackage{times}
\usepackage{latexsym}

\usepackage[T1]{fontenc}

\usepackage[utf8]{inputenc}

\usepackage{microtype}

\usepackage{inconsolata}

\usepackage{graphicx}
\usepackage{amsmath}
\usepackage{amsfonts}
\usepackage{multirow}
\usepackage{amsthm}
\usepackage{booktabs}
\usepackage{epstopdf}
\usepackage{xcolor}
\usepackage[labelformat=simple]{subcaption}
\usepackage{caption}
\usepackage{bbm}
\usepackage{dsfont}
\usepackage[ruled,linesnumbered]{algorithm2e}
\usepackage{setspace}
\usepackage{braket}
\usepackage{tablefootnote}
\usepackage{subcaption}
\usepackage[normalem]{ulem}
\usepackage{cleveref}
\usepackage{makecell}
\usepackage{wrapfig}
\usepackage{tcolorbox}
\usepackage{footnote}
\title{Empowering GraphRAG with Knowledge Filtering and Integration}

\author{Kai Guo$^{1}$, Harry Shomer$^1$, Shenglai Zeng$^1$, Haoyu Han$^1$, Yu Wang$^{2}$, Jiliang Tang$^{1}$ 
\\ 
$^1$Michigan State University  \quad $^2$ University of Oregon  
  \\
\{guokai1, shomerha, zengshe1, hanhaoy1, tangjili\}@msu.edu, \\
\{yuwang\}@uoregon.edu
}

\begin{document}
\maketitle
\begin{abstract}
In recent years, large language models (LLMs) have revolutionized the field of natural language processing. However, they often suffer from knowledge gaps and hallucinations. Graph retrieval-augmented generation (GraphRAG) enhances LLM reasoning by integrating structured knowledge from external graphs. However, we identify two key challenges that plague GraphRAG: (1) Retrieving noisy and irrelevant information can degrade performance and (2) Excessive reliance on external knowledge suppresses the model’s intrinsic reasoning.  
To address these issues, we propose GraphRAG-FI (Filtering \& Integration), consisting of GraphRAG-Filtering and GraphRAG-Integration. GraphRAG-Filtering employs a two-stage filtering mechanism to refine retrieved information. GraphRAG-Integration employs a logits-based selection strategy to balance external knowledge from GraphRAG with the LLM’s intrinsic reasoning, reducing over-reliance on retrievals.
Experiments on knowledge graph QA tasks demonstrate that GraphRAG-FI significantly improves reasoning performance across multiple backbone models, establishing a more reliable and effective GraphRAG framework.
\end{abstract}

\section{Introduction}

Large language models (LLMs) have achieved remarkable success in NLP tasks, particularly in tasks that require complex reasoning~\cite{havrillaglore,wu2023symbol,hao2023reasoning}. However, despite their strengths, LLMs are prone to hallucinations, resulting in incorrect or poor reasoning~\cite{ji2023towards,huang2024survey,sriramanan2025llm}. GraphRAG techniques have emerged as a promising solution to this problem~\cite{han2024retrieval,zhang2025survey,he2025g,mavromatis2024gnn}, by integrating relevant information from external graphs. Knowledge graphs, which store facts in the form of a graph, are commonly used for this problem. Specifically, relevant facts (i.e., triples) or paths are extracted from the knowledge graph and used to enrich the context of the LLMs with structured and reliable information~\cite{luo2024rog,li2024subgraphrag,ma2024think}. This approach has shown ability to improve the reasoning capabilities and reduce the presence of hallucinations in LLMs~\cite{sunthink,li2024subgraphrag,dong2024don}. 


To better assess the efficacy of GraphRAG, in Section~\ref{sec: Preliminary} we conduct a preliminary study comparing its performance with an LLM-only model (i.e., LLM without GraphRAG). This comparison reveals both the advantages and limitations of GraphRAG. While GraphRAG improved reasoning accuracy by correcting some LLM errors, it also introduces some notable weaknesses. For example, incorporating external knowledge will sometimes cause questions that were originally answered correctly by the LLM to be misclassified.  This highlights the dangers of retrieving irrelevant information.  Furthermore, excessive retrieval compounds this issue by introducing both noise and redundant information, thus further hindering the reasoning process. 

Meanwhile, we find that {\it LLM-only and GraphRAG can complement one another}. Specifically, GraphRAG can enhance reasoning for those questions LLMs lack knowledge of; while excessive reliance on external information may cause the model to overlook internally known correct answers. These findings highlight two key limitations of existing GraphRAG methods. {\bf First}, GraphRAG is highly susceptible to retrieving irrelevant or misleading information. {\bf Second}, GraphRAG struggles to balance external retrieval with the LLM’s internal knowledge, 
often missing parts of the answer that the LLM-only model can provide using its own knowledge.




Inspired by these findings, we propose a novel design that aims to address these issues. First, we aim to enhance the retrieval quality to better avoid retrieving irrelevant information. Second, we integrate GraphRAG with an LLM’s intrinsic reasoning ability, thus only using GraphRAG when external knowledge is necessary. In particular, to mitigate the issue of retrieving irrelevant information, we introduce a two-stage filtering process. 
Furthermore, to mitigate GraphRAG from over-relying on retrieved information while underutilizing the LLM’s inherent reasoning ability, we introduce a logits-based selection mechanism that dynamically integrates LLMs' standalone answers with GraphRAG’s outputs. This approach ensures that the final response effectively balances external knowledge with the model’s internal reasoning. The main contributions of our work are summarized as follows:  

\begin{itemize}  

\item We identify two key challenges in GraphRAG: {\bf (1)} It is susceptible to errors by retrieving irrelevant or misleading information. {\bf (2)} It overemphasizes the externally retrieved knowledge, at the expense of the intrinsic reasoning capabilities of LLMs. 

\item We introduce a novel approach that enhances GraphRAG by incorporating a two-stage filtering mechanism to refine the retrieved knowledge and dynamically integrate this knowledge with a LLMs' standalone reasoning capabilities.

\item Extensive experiments on knowledge graph QA demonstrate the effectiveness of our method across multiple backbone models.  

\end{itemize}
\section{Related work}
\textbf{GraphRAG.} 
GraphRAG aims to address hallucinations and outdated knowledge in LLMs by incorporating additional information retrieved from external knowledge bases~\cite{sunthink,li2024subgraphrag,dong2024don}.  G-Retriever~\cite{he2025g} identifies relevant nodes and edges for a given query based on cosine similarity, and then constructs a subgraph to aid in the generation process. Similarly, RoG~\cite{luo2024rog} introduces a planning-retrieval-reasoning framework, where it retrieves reasoning paths guided by a planning module and performs reasoning using these paths. On the other hand, GNN-RAG~\cite{mavromatis2024gnn} leverages Graph Neural Networks (GNNs)~\cite{kipf2016semi} to process the intricate graph structures within knowledge graphs, enabling effective retrieval. They also use retrieval augmentation techniques to enhance diversity.
However, the effectiveness of these methods is heavily dependent on the quality of the retrieved information, and their performance significantly declines when the retrieved graph data is either noisy or unrelated to the query~\cite{he2025g} .

\textbf{Filter Methods.}
Filtering attempts to only keep those pieces of retrieved information that are relevant to the given query~\cite{gao2025frag}.
ChunkRAG~\cite{singh2024chunkrag} tries to improve RAG systems by assessing and filtering retrieved data at the chunk level, with each "chunk" representing a concise and coherent segment of a document. This method first applies semantic chunking to partition documents into meaningful sections. It then leverages LLM-based relevance scoring to evaluate how well each chunk aligns with the user query.
\citet{zeng2024towards} thoroughly investigate LLM representation behaviors in relation to RAG, uncovering distinct patterns between positive and negative samples in the representation space. This distinction enables representation-based methods to achieve significantly better performance for certain tasks. Building on these insights, they introduce Rep-PCA, which employs representation classifiers for knowledge filtering.
RoK~\cite{wang2024reasoning} refines the reasoning paths within the subgraph by computing the average PageRank score for each path. Similarly, \citet{he2024g} use PageRank to identify the most relevant entities.

\section{Preliminary studies}
\label{sec: Preliminary}
To evaluate the effectiveness of GraphRAG, we compare the performance with and without retrieved external knowledge. Furthermore, we analyze the attention scores of the LLM to assess its ability to discern both the relevance and importance of the retrieved information. Lastly, we evaluate the performance of internal knowledge filtering.

\subsection{Experimental settings}


In this section, we aim to study the importance of retrieving external information when using GraphRAG for knowledge graph QA. To do so, we report the QA performance when using: LLM {\it with} GraphRAG and LLM {\it w/o} GraphRAG (i.e., LLM-only).
For GraphRAG, we use RoG~\cite{luo2024rog} and GNN-RAG~\cite{mavromatis2024gnn}. 
For the LLM-only experiments, we use the fine-tuned LLaMA 2-7B model, which is the same LLM used by RoG. The experiments are conducted on two common datasets the WebQSP~\cite{yih2016value} and CWQ~\cite{talmor2018web} datasets. In this study, we mainly use the F1 score to evaluate the performance.


\begin{figure}     \centering     
\includegraphics[scale=0.2]{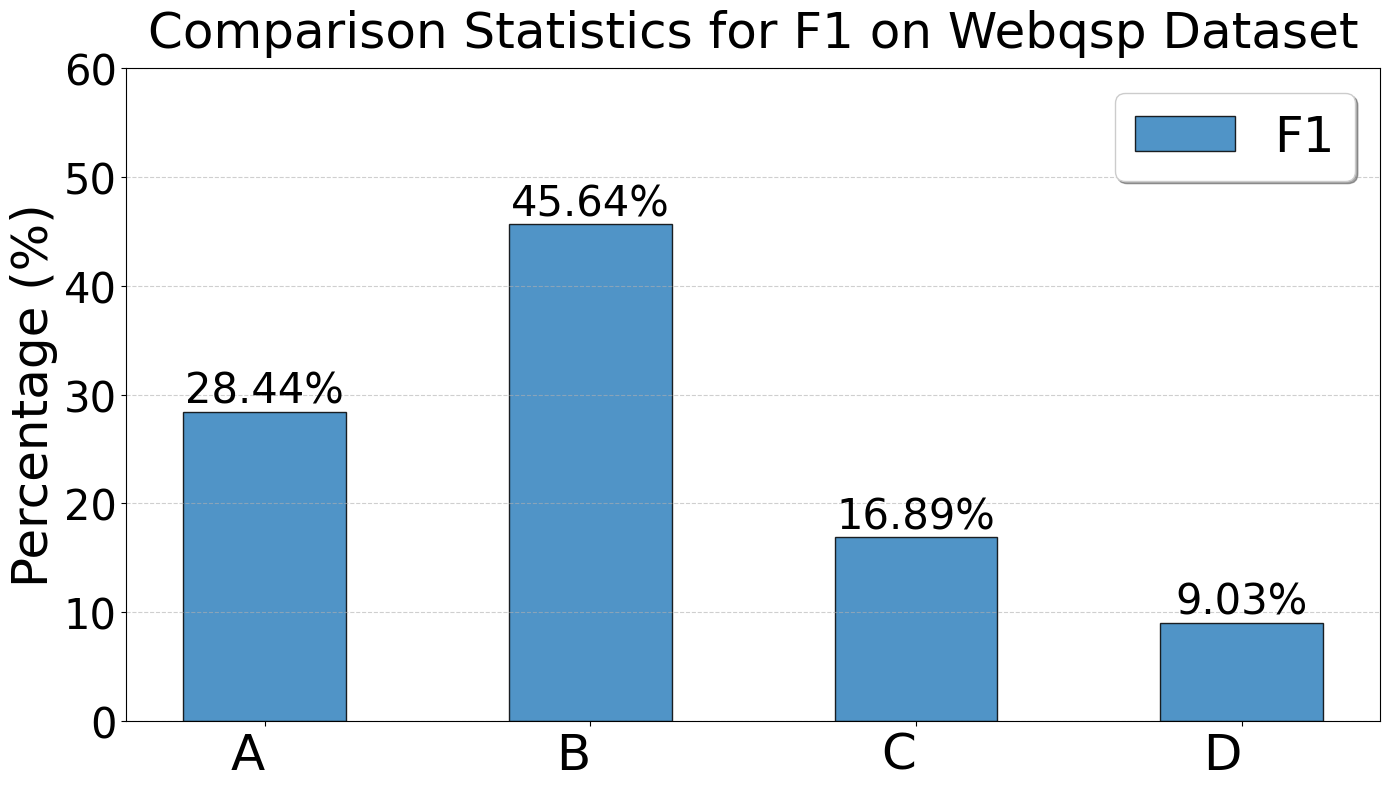}  
\captionsetup{skip=10pt}
\caption{Category A includes cases where both GraphRAG and the LLM-only model are correct. Category B covers instances where GraphRAG outperforms the LLM-only model, while Category C includes cases where the LLM-only model performs better than GraphRAG. Category D represents cases where both models fail.}   \label{fig: overlap} 
\end{figure}


\begin{figure}[ht]    \centering     
\includegraphics[scale=0.22]{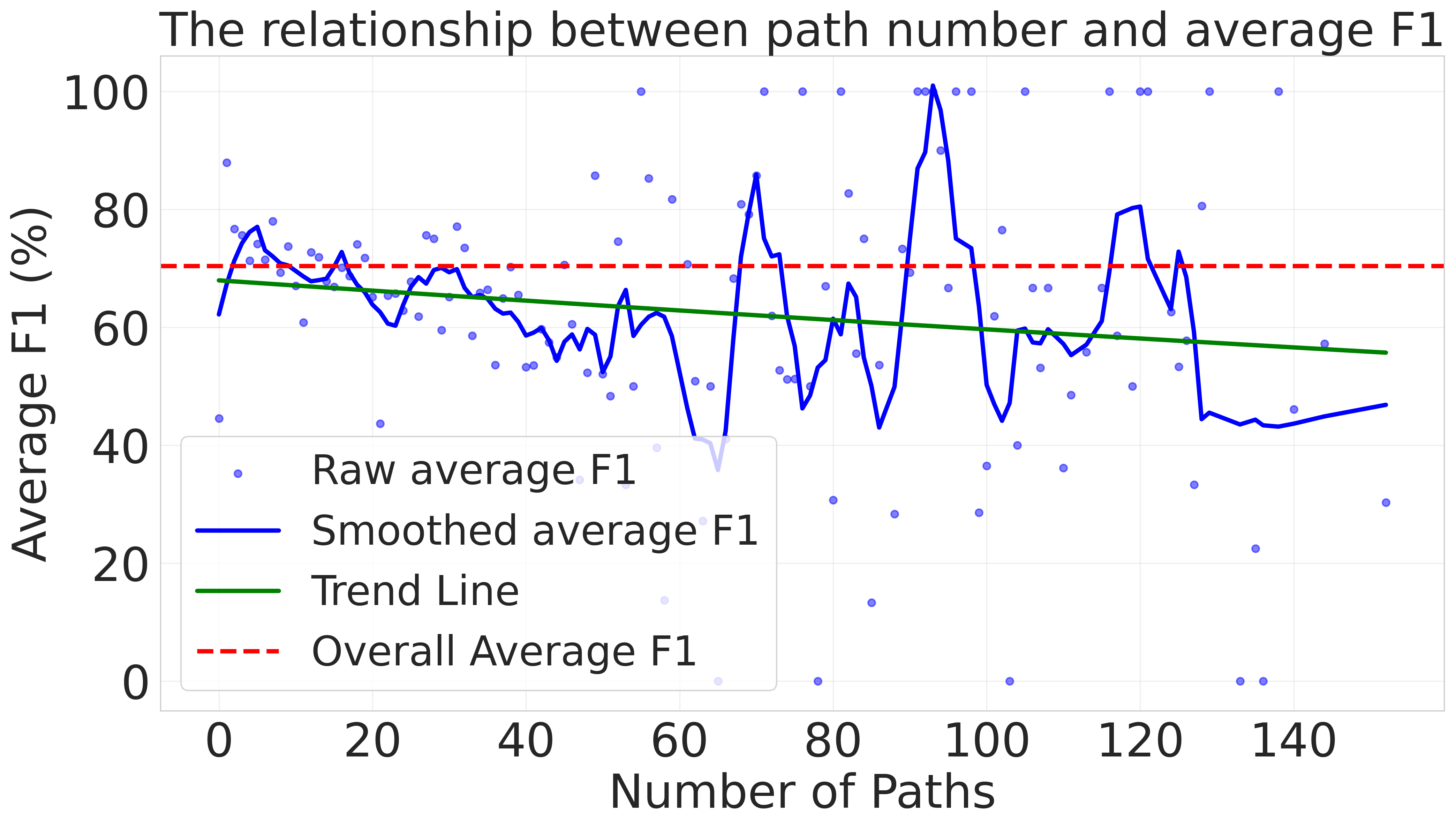}  
\captionsetup{skip=10pt}
\caption{The relationship between path number and average F1}   \label{fig: num} 
\end{figure}

\subsection{The Impact of GraphRAG} \label{sec:graphrag_impact}



To understanding the effectiveness of GraphRAG, we compare prediction outcomes between LLM {\it with} GraphRAG and LLM {\it w/o} GraphRAG (i.e., LLM-only). 
We categorize the results into four groups based on F1 scores, as shown in the Figure~\ref{fig: overlap}. Category A includes cases where both GraphRAG and the LLM-only model provide correct answers. Category B consists of instances where GraphRAG produces a more accurate answer than the LLM-only model. Category C includes cases where the LLM-only model outperforms GraphRAG. Finally, Category D represents instances where both GraphRAG and the LLM-only model fail to generate the correct answer.
Figure~\ref{fig: overlap} illustrates the key observations from our experiments. While GraphRAG enhances certain predictions, it also introduces notable challenges that require further investigation.

\paragraph{Positive Impact of GraphRAG}  
GraphRAG can enhance the LLM's reasoning capabilities by correcting errors that the standalone model would typically commit. Notably, in the category \textbf{B}, 45.64\% of previously incorrect responses were successfully rectified with the integration of GraphRAG. This highlights the advantage of leveraging structured knowledge graphs to boost LLM performance.

\paragraph{Limited Impact of GraphRAG} Category {\bf A} contains those answers where both GraphRAG and LLM-only are correct. This show that GraphRAG can sometimes preserve the performance of a LLM when the LLM already possesses the correct knowledge. Conversely, category \textbf{D}, representing 9.03\% of cases, corresponds to those cases where GraphRAG fails to enhance the model's accuracy. For this category, neither the standalone LLM nor GraphRAG are able to provide the correct answer. This pattern implies that GraphRAG does not always access or incorporate sufficiently informative or relevant knowledge.

\paragraph{Negative Impact of GraphRAG}  
A notable drawback of GraphRAG is that will occasionally 
degrade the performance of a standalone LLM. That is, it will sometimes lead to wrong predictions for queries that the standalone LLM originally got right. These instances are represented by category {\bf C} and accounts for 16.89\% of samples when evaluating via the F1 score. 
In these cases, GraphRAG misleads the model rather than improving it. This suggests that some of the retrieved information may be incorrect, noisy, or irrelevant, ultimately leading to poorer predictions. Therefore, in some cases, LLMs without GraphRAG outperform those with GraphRAG,  because existing works have shown that LLMs tend to over-rely on external information~\cite{ren2023investigating,tan2024blinded,wang2023self,ni2024llms,zeng2024good}. When retrieval is insufficient or the quality of retrieved knowledge is low, this reliance can degrade generation quality.

\subsection{The Impact of the Number of Retrieved Paths}

Due to the structure of knowledge graphs, nodes with high degrees and numerous relational edges have a greater likelihood of yielding a large number of retrieved paths.
In this subsection, we study the impact of the number of retrieved paths on performance.
Figure~\ref{fig: num} illustrates the relationship between the number of retrieved paths and the model’s performance. Interestingly, as indicated by the smoothed line (blue), incorporating a moderate amount of retrieved information enhances performance. However, increasing the number of retrieved paths ultimately leads to a decline in performance. This trend (green line) suggests that retrieving too much information will introduce noise, making it harder for the model to use the correct and relevant knowledge for the task. This phenomenon thus highlights an important insight -- {\bf more information does not necessarily indicate better performance}. Instead, an overabundance of retrieved data can overwhelm the model with irrelevant details. 
This observation underscores the necessity for effective filtering mechanisms that can prioritize high-quality, relevant knowledge while discarding extraneous or misleading information.


\subsection{Attention Reflects the Importance of Retrieved Information}
\begin{figure}[ht]     \centering     
\includegraphics[scale=0.22]{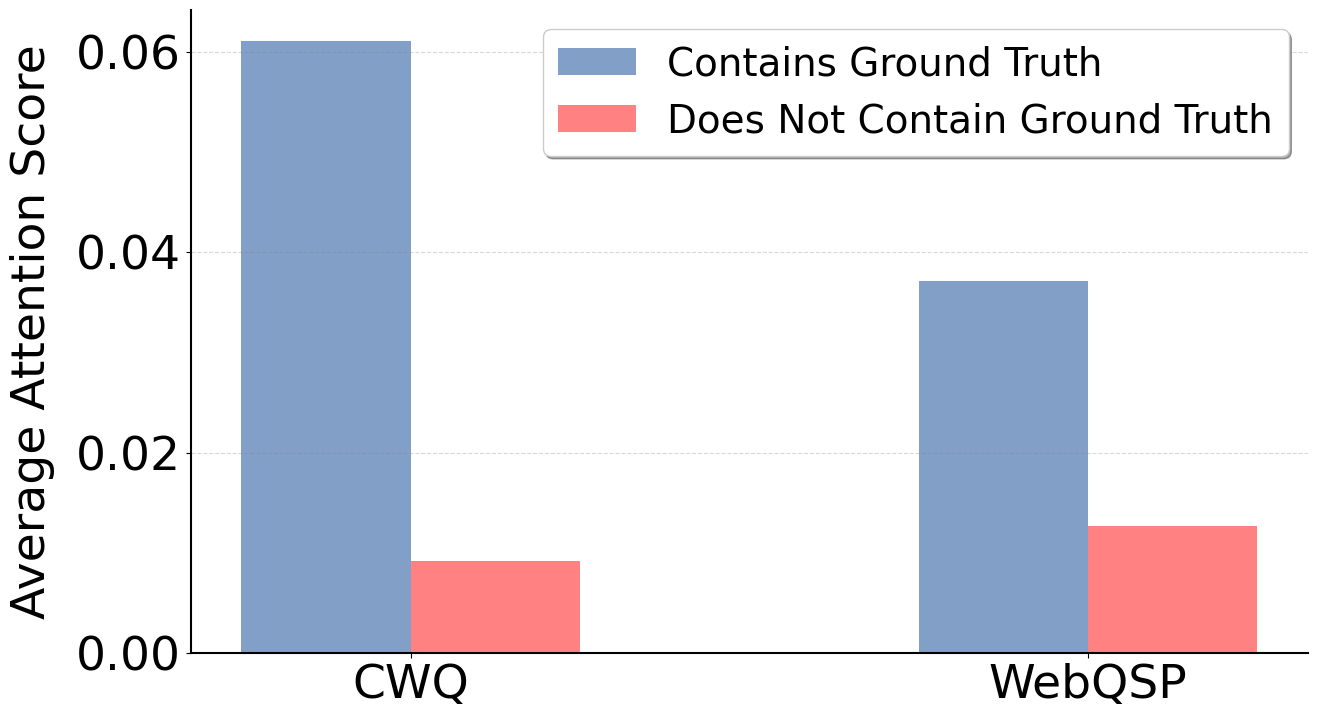}  
\captionsetup{skip=10pt}
\caption{Attention Scores for Retrieved Information With/Without Ground Truth}   \label{fig: attention} 
\end{figure}

In this subsection, we  analyze the ability of the LLM to distinguish the importance of retrieved external knowledge. 
The attention scores of a LLM can provide a natural indicator of the relevance and significance of the retrieved knowledge~\cite{yang2024tidaldecode,ben2024attend}. The attention scores, derived from the model’s internal mechanisms, effectively capture which pieces of information are most influential in reaching the final decision. Inspired by recent work~\cite{chuang2023dola,halawi2023overthinking}, which suggests that attention scores in the middle layers are more effective.  We examine the attention scores of the (middle + 2)-th layer in the LLM for each retrieved path.
We obtain the attention scores for all retrieved paths and categorize them into two groups: (1) paths that contain the ground truth and (2) paths that do not. We then compute the average attention score for each group and present the results in Figure~\ref{fig: attention}. As demonstrated in Figure~\ref{fig: attention}, there is a clear alignment between the attention scores and the ground truth labels, suggesting that these scores can be used to assess the relevance of retrieved information.

This observation inspires a key insight: 
The attention scores highlight the most significant retrieved information, suggesting their potential use in filtering out noisy or irrelevant knowledge. Since retrieved information with lower attention scores contribute minimally to the final output, they can be pruned to streamline retrieval and enhance overall performance. 

\subsection{Internal Knowledge Filtering}  \label{sec:llm_logits}
Large language models (LLMs) generate responses that may contain both correct and incorrect information. To assess the reliability of these responses, we analyze the associated logits, which represent the model's confidence in its predictions. Typically, higher confidence correlates with correctness~\cite{ma2025estimating,virk2024enhancing}. Leveraging this property, we implement ``Internal Knowledge Filtering'', which uses the logits to help refine the answer selection.The logits of answer can be directly obtained from the LLM's output.
Formally, let \(A_L\) denote the sets of answer candidates from  the LLM model. Furthermore, let it's corresponding logits after softmax function be given by \(\ell_L(a)\). The filtering step is given by the following:
\begin{align} 
    &A_L^{\text{filtered}} = \{ a \in A_L \mid \ell_L(a) \geq \tau_L \},
\end{align}
where \(\tau_L=1\).  This allows us to filter out the responses that the LLM has low-confidence in. 
The experimental results are shown in Table~\ref{fig: logits}. We can clearly see that that leveraging logits to filter out low-confidence responses has a large positive effect on performance. In this way, we can reconsider intrinsic knowledge and apply this approach to GraphRAG to better balance internal and external knowledge base on logits.

\begin{table}[h]
    \centering
\caption{Impact of logits on LLM performance}   \label{fig: logits}
\resizebox{0.45\textwidth}{!}{
    \begin{tabular}{lcc|cc}
        \toprule
        Methods & \multicolumn{2}{c|}{WebQSP} & \multicolumn{2}{c}{CWQ} \\
        \cmidrule(lr){2-5}
        & Hit & F1 & Hit & F1 \\
        \midrule
        LLM & 66.15 & 49.97 & 40.27 & 34.17 \\
        LLM with Logits & 84.17 & 76.74 & 61.83 & 58.19 \\
        \bottomrule
    \end{tabular}}
\end{table}

\subsection{Discussions}

In this subsection, we summarize the key findings and discussions from our preliminary study. The performance issues observed in GraphRAG primarily arise from two key factors. (1) \textbf{Noisy or Irrelevant Retrieval:} Some retrieved paths contain irrelevant or misleading information. This negatively impacts the model's ability to properly answer the query. Furthermore, this noise can introduce conflicting or unnecessary information that hinders the decision-making process rather than improving it. (2) \textbf{Lack of Consideration for LLM’s Own Knowledge:} GraphRAG does not always take into account the inherent reasoning ability of the LLM itself. In some cases, the retrieved information overrides the LLM’s correct predictions, leading to performance degradation rather than enhancement. A more adaptive approach is needed to balance external knowledge retrieval with the model’s internal knowledge.

\section{Method}
\begin{figure*}     \centering     
\includegraphics[scale=0.5]{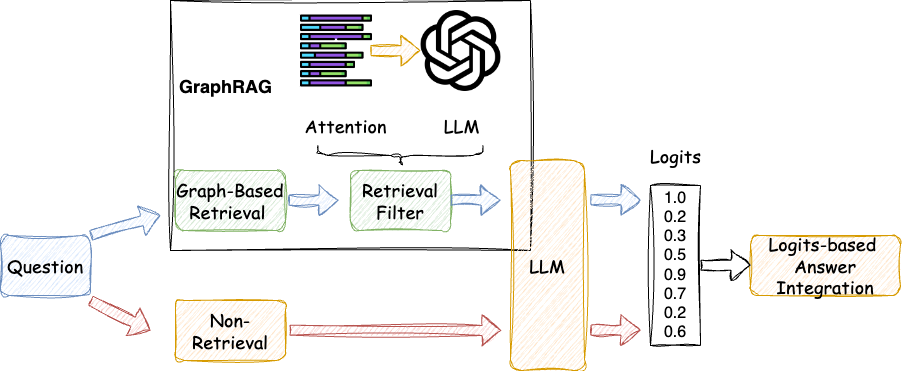}  
\captionsetup{skip=10pt}
\caption{An overview of the GraphRAG-FI framework. }   \label{fig: model} 
\end{figure*}

Based on our analysis, we propose a new framework to address the identified challenges, guided by two key insights:  
{\bf (1)} {\it Filtering retrieved information}: Given the tendency of GraphRAG to retrieve irrelevant or incorrect retrieved information, it is essential to refine the retrieved knowledge. {\bf (2)} {\it Properly leveraging the LLMs standalone capabilities}: The LLM itself can often correctly answer some questions. It's thus necessary to effectively integrate and use the inherent reasoning ability of LLMs along with GraphRAG.  

An overview of our framework {\bf GraphRAG-FI} is given in Figure~\ref{fig: model}. It consists of two core components: {\bf GraphRAG-Filtering} and {\bf GraphRAG-Integration}. GraphRAG-Filtering first refines the retrieved information by removing irrelevant or misleading knowledge. GraphRAG-Integration module balances the retrieved knowledge with the LLM’s inherent reasoning ability, thereby mitigating the overuse of retrieved information that can negatively impact performance. In the following subsections, we will introduce each component of our framework in detail.


\subsection{GraphRAG-Filtering}
Let \( P = \{p_1, p_2, \ldots, p_N\} \) denote the set of $N$ retrieved paths or triplets, where each path \( p_i \) is assigned an attention score \( a_i \). 
Then we design filtering via the following two stages.

\paragraph{Stage 1: Coarse Filtering using Attention:}  
In the first stage, we perform a coarse filtering by retaining only those paths whose attention scores exceeds a  threshold \( \tau \). This is given formally by:
\begin{equation}
P_{\text{coarse}} = \{ p_i \in P \mid a_i \geq \tau \}.
\end{equation}

\paragraph{Stage 2: Fine Filtering via LLMs:}  
After the initial coarse filtering, which significantly reduces the number of candidate paths, we perform a more precise evaluation with  a LLM on the remaining subset. This two-stage filtering approach not only enhances the quality of the retrieved paths but also greatly reduces the overall cost by limiting the use of the LLM to only those paths deemed promising in the first stage. Let \( f(p) \) represent the evaluation score provided by the LLM for a path \( p \), and let \( \tau' \) be the corresponding threshold. The final set of filtered paths is then given by:
\begin{equation}
P_{\text{final}} = \{ p \in P_{\text{coarse}} \mid f(p) \geq \tau' \},
\end{equation}
where \( P_{\text{coarse}} \) is the set of paths that passed the coarse filtering stage, $\tau'$ is not predefined but is determined by the LLM itself.

\paragraph{Prompt Construction:}  

After the two filtering stages, we incorporate the selected paths and query into the prompt to further guide the model's reasoning. The prompt contains the following two types of retrieved paths:
\begin{itemize}
    \item {\bf High Priority Paths}: These are the final filtered paths given by $P_{\text{final}}$, which are considered the most reliable.  
    \item {\bf Additional Paths}: We also consider the the remaining paths included by the coarse filter but removed via the fine filter, $P_{\text{coarse}} - P_{\text{final}}$.  We conjecture that while they may not be as important as those paths in  $P_{\text{final}}$, they can still offer some useful supplementary context.
\end{itemize}

The new prompt is then constructed by first inserting a header for the high-priority paths, followed by each path on a separate line. The same process is repeated for the additional paths. By structuring the prompt in this way, we are able to clearly delineate the paths by their priority. This ensures that the most critical information ($P_{\text{final}}$) is emphasized and processed first, while still incorporating the supplementary context from the additional paths. An example prompt is given in Appendix~\ref{sec:prompt}.

\subsection{Integration with LLMs' Internal Knowledge}
As noted in Section~\ref{sec:graphrag_impact}, in addition to ensuring we only retrieve high-quality information, we also want to retain internal knowledge of the LLMs. As such, we want to also integrate the capabilities of just the LLM into our framework. However, a challenge is knowing when to defer to which method. {\it When do we trust the answers given by GraphRAG and when the standalone LLM}? Furthermore, {\it how do we fuse the answers given by both methods}? 

To achieve this goal, we need a method to determine which answers produced by both LLM-only and GraphRAG are actually relevant. In Section~\ref{sec:llm_logits}, we found that the LLM's logits can provide a useful tool to refine the potential answers. That is, focusing only on those answers that are given a higher confidence is helpful. This naturally provides us with an easy way to focus on just the high-quality information. For both GraphRAG and the LLM-only model, we filter the answers based on their logits, ensuring that only high-confidence responses are retained. After this logits-based filtering, the refined answers from both sources are combined to produce the final answer, thereby enhancing robustness and accuracy.

Formally, let \(A_G\) and \(A_L\) denote the sets of answer candidates from GraphRAG and the LLM-only model, respectively. We further use $a$ to indicate a single candidate answer in either set. Furthermore, let their corresponding logits after the softmax function be given by \(\ell_G(a)\) and \(\ell_L(a)\). The filtering step is given by the following:
\begin{align} 
    &A_G^{\text{filtered}} = \{ a \in A_G \mid \ell_G(a) \geq \tau_G \}, \\
    &A_L^{\text{filtered}} = \{ a \in A_L \mid \ell_L(a) \geq \tau_L \},
\end{align}
where \(\tau_G\) and \(\tau_L\) are predefined thresholds, $\tau_L$ is set to 1. Subsequently, the final answer is determined by combining the filtered sets:
\begin{equation}
A_{\text{final}} = \operatorname{Combine}\Bigl( A_G^{\text{filtered}},\, A_L^{\text{filtered}} \Bigr),
\end{equation}
where \(\operatorname{Combine}(\cdot)\) denotes the function that integrates the filtered answers into the final reliable output. 


\begin{table*}[ht]
    \centering
    \caption{Performance comparison with different baselines on the two KGQA datasets.}
    \label{tab: all_results}
\begin{tabular}{c|llccc}
\hline
\multirow{2}{*}{Type} & \multirow{2}{*}{Methods} & \multicolumn{2}{c}{WebQSP} & \multicolumn{2}{c}{CWQ} \\ \cline{3-6} 
 &  & Hit & F1 & Hit & F1 \\ \hline
\multirow{5}{*}{LLMs} & Flan-T5-xl\citep{chung2024scaling} & 31.0 & - & 14.7 & - \\
 & Alpaca-7B\citep{taori2023stanford} & 51.8 & - & 27.4 & - \\
 & LLaMA2-Chat-7B\citep{touvron2023llama} & 64.4 & - & 34.6 & - \\
 & ChatGPT & 66.8 & - & 39.9 & - \\
 & ChatGPT+CoT & 75.6 & - & 48.9 & - \\ \hline
\multirow{15}{*}{LLMs+KGs}
 & ROG & 86.73 & 70.75 & 61.91 & 54.95 \\
  & ROG + Similarity & 85.50 & 69.38 & 61.62 & 54.38\\
  & ROG + PageRank & 85.44 & 69.60 & 61.34 & 54.41\\
  & ROG + GraphRAG-Filtering & 87.40 & 73.41 & 63.86 & \textbf{57.25}\\
  & ROG + GraphRAG-FI & \textbf{89.25} & \textbf{73.86} & \textbf{64.82} & 55.12\\ \cmidrule(lr){2-6} 
 & GNN-RAG & 90.11 & 73.25 & 69.10 & 60.55 \\
  & GNN-RAG + Similarity & 89.68 & 72.17 & 68.50 & 60.26\\
  & GNN-RAG + PageRank & 89.18 & 71.92 & 66.75 & 58.73\\
  & GNN-RAG + GraphRAG-Filtering & 91.28 & 74.74 & 69.70 & \textbf{60.96}\\
  & GNN-RAG + GraphRAG-FI & \textbf{91.89} & \textbf{75.98} & \textbf{71.12} & 60.34\\ \cmidrule(lr){2-6}
 & SubgraphRAG & 76.90 & 64.65 & 53.87 & 50.43\\
  & SubgraphRAG + Similarity & 72.72 & 59.98 & 52.05 & 48.27\\
  & SubgraphRAG + PageRank & 61.79 & 50.65 &46.75 & 43.23\\
  & SubgraphRAG + GraphRAG-Filtering & 81.01 & \textbf{68.40} & 58.82 & \textbf{54.71}\\
  & SubgraphRAG + GraphRAG-FI & \textbf{81.08} & 68.28 & \textbf{58.96} & 52.52\\ \cline{2-6}
 \hline
\end{tabular}
\end{table*}
\section{Experiment}
In our experiments, we seek to address the following research questions:  
\textbf{RQ1:} How effective is the proposed method when applied to state-of-the-art GraphRAG retrievers in the knowledge graph QA task?  
\textbf{RQ2:} How does the proposed method compare to other filtering approaches?
\textbf{RQ3:} How does the performance change when more noisy information is introduced? and 
\textbf{RQ4:} What is the impact of the two modules on performance?

\subsection{Experiment Settings}
\textbf{Datasets.} To assess the effectiveness of our method, we evaluate it on two widely recognized KGQA benchmark datasets: WebQSP~\cite{yih2016value} and CWQ~\cite{talmor2018web}. WebQSP contains 4,737 natural language questions that require reasoning over paths of up to two hops. In contrast, CWQ includes 34,699 more complex questions that necessitate multi-hop reasoning over up to four hops. Both datasets are built upon Freebase~, which consists of around 88 million entities, 20 thousand relations, and 126 million triples. Further details on the datasets are provided in Appendix~\ref{sec:dataset}.

\textbf{Retriever Backbones.}
Our framework adopts three existing retrieval methods as its backbone: path-based retrieval (ROG~\cite{luo2024rog}), GNN~\cite{mavromatis2024gnn}), and subgraph-based retrieval (SubgraphRAG~\cite{li2024subgraphrag}). Path-based retrieval extracts relevant paths using heuristics or shortest-path algorithms, while GNN-based retrieval leverages a Graph Neural Network to learn and retrieve informative paths. In contrast, subgraph-based retrieval retrieves relevant subgraphs and encodes them as triples, enabling fine-grained relational reasoning. Therefore, both path-based and GNN-based methods generate paths as input for the LLM. Lastly, subgraph-based methods give triples (i.e., edges) as input to the LLM that take the form of \((h, r, t)\). By considering these three methods, we are able to test our framework on a diverse set of retrieval methods.

\textbf{Filter Baselines.}
The most commonly used filtering methods for RAG are similarity-based approaches used in~\cite{gao2025frag}. Similarity-based methods evaluate the relevance of retrieved information by measuring feature similarity. For retrieval over graphs, PageRank-based filtering is widely adopted~\cite{wang2024reasoning}. PageRank-based filtering leverages the graph structure to rank nodes based on their connectivity and importance. These methods provide a baseline filtering mechanism for refining the retrieved results.

\textbf{Implementation and Evaluation Metrics.}
We use LLaMA2-Chat-7B from ROG as the LLM backbone, which is instruction-finetuned on the training split of WebQSP and CWQ, as well as Freebase, for three epochs. For the similarity-based filter, we utilize SentenceTransformer (`all-MiniLM-L6-v2') to generate representations for retrieval.
We evaluate our retrieval methods using both Hit Rate (Hit) and F1-score (F1). 
Hit Rate measures the proportion of relevant items successfully retrieved, reflecting retrieval effectiveness. F1-score balances precision and recall, providing a comprehensive assessment of retrieval quality. These metrics ensure a robust evaluation of retrieval performance.
We adjust the thresholds \(\tau\) and \(\tau_G\) within the ranges [top 40, top 50] and [0.4, 0.5], respectively.
\subsection{Main Results}
In this section, we evaluate the performance of our method with various retrievers and compare it against baseline filter models.

\noindent \textbf{RQ1: KGQA Performance Comparison}.
In this subsection, we apply our method to different retrievers, including the path-based retriever, GNN-based retriever, and subgraph-based retriever. The results presented in Table~\ref{tab: all_results} demonstrate that our method consistently improves all retrievers, achieving an average improvement of 3.81\% in Hit and 2.35\% in F1 over ROG, 2.46\% in Hit and 1.7\% in F1 over GNN-RAG, and significant gains of 7.47\% in Hit and 4.88\% in F1 over SubgraphRAG across two datasets.
These results demonstrate that our approach is effective across different retrieval paradigms, reinforcing its adaptability to various retrieval strategies in QA tasks.

\noindent {\bf RQ2: Comparison with other filter methods.} 
We compare our method against other filtering baselines, with the results presented in Table~\ref{tab: all_results}. Our approach consistently outperforms competing methods across both datasets and retriever types. Specifically, for ROG, our method can achieve an average improvement of 4.78\% in Hit and 3.95\% in F1 compared to similarity-based filtering on both datasets. Furthermore, compared to the PageRank-based filtering method, our approach yields an average increase of 5.03\% in Hit and 3.70\% in F1 across both datasets. These results highlight the superiority of our method in enhancing retrieval effectiveness and overall performance.

\begin{table}[h]
    \centering

\caption{Performance when adding more noise}   \label{tab: noise}

\resizebox{0.45\textwidth}{!}{
    \begin{tabular}{lcc|cc}
        \toprule
        Methods & \multicolumn{2}{c|}{WebQSP} & \multicolumn{2}{c}{CWQ} \\
        \cmidrule(lr){2-5}
        & Hit & F1 & Hit & F1 \\
        \midrule
        ROG-original & 86.73 & 70.75 & 61.91 & 54.95 \\
        ROG* & 85.87 & 68.81 & 60.49 & 53.72 \\
        ROG* + GraphRAG-Filtering & 86.61 & 73.01 & 61.91 & 55.67 \\
        \bottomrule
    \end{tabular}}
\end{table}

\subsection{Robustness to Noise }
In this subsection, we evaluate robustness of different methods to noise.  To evaluate the noise resistance of the backbone model and our filter method, we use GPT to generate 30 additional noise paths that contain both irrelevant and incorrect information. This information is then incorporated into the retrieved context. We then analyze the impact of this noise on performance. The experimental results presented in Table~\ref{tab: noise}, ROG* represents the cases where noise is introduced. As the noise level increases, the Hit score decreases by 2.29\%, and the F1 score drops by 2.23\% on the CWQ dataset, highlighting the model's sensitivity to noise. However, when applying our method, we observe a 2.23\% improvement in Hit and a 3.63\% improvement in F1 over ROG* on CWQ. These results demonstrate the effectiveness of our approach in mitigating the negative impact of noisy retrieval.

\subsection{Ablation Study}
\label{sec:ablation}
We conduct an ablation study to analyze the effectiveness of the filtering module and integrating module in GraphRAG-FI. From the results in Table~\ref{tab: ablation}, we can see that GraphRAG-Filtering is useful for the ROG retriever, as it improves both the F1 and Hit scores. For example, GraphRAG-Filtering increases the F1 score by 4.19\%  and the Hit score by 3.15\% on CWQ dataset. We also see a boost in performance for GraphRAG-Integration, with a 1.60\% and 2.62\% increase in F1 and Hit score, respectively, on WebQSP. These results demonstrate the effectiveness of our two components.
\begin{table}[h]
    \centering
\caption{Ablation study.}   \label{tab: ablation}

\resizebox{0.48\textwidth}{!}{
    \begin{tabular}{lcc|cc}
        \toprule
        Methods & \multicolumn{2}{c|}{WebQSP} & \multicolumn{2}{c}{CWQ} \\
        \cmidrule(lr){2-5}
        & Hit & F1 & Hit & F1 \\
        \midrule
        ROG-original & 86.73 & 70.75 & 61.91 & 54.95 \\
        ROG + GraphRAG-Filtering & 87.40 & 73.41 & 63.86 & \textbf{57.25} \\
        ROG + GraphRAG-Integration & 89.00 & 71.88 & 64.25 & 55.19 \\
         ROG + GraphRAG-FI & \textbf{89.25} & \textbf{73.86} & \textbf{64.82} & 55.12 \\

        \bottomrule
    \end{tabular}}
\end{table}

\section{Conclusion}
In this work, we propose GraphRAG-FI (Filtering \& Integration), an enhanced GraphRAG framework that addresses key challenges in graph retrieval-augmented generation. By incorporating GraphRAG-Filtering, which utilizes a two-stage filtering mechanism to refine retrieved information, and GraphRAG-Integration, which employs a logits-based selection strategy to balance retrieval and intrinsic reasoning, our approach mitigates the impact of noisy retrievals and excessive dependence on external knowledge. Experimental results on knowledge graph QA tasks demonstrate that GraphRAG-FI significantly improves reasoning accuracy across multiple backbone models, establishing a more reliable and effective GraphRAG framework.
\clearpage
\section*{Limitations}
In this work, we identify two key challenges in GraphRAG: (1) it is prone to errors due to the retrieval of irrelevant or misleading information, and (2) it places excessive emphasis on externally retrieved knowledge, which can diminish the intrinsic reasoning capabilities of LLMs. Future research will first explore a broader range of large language models to evaluate their effectiveness within GraphRAG. Additionally, further investigation into diverse filtering methods could enhance the refinement of retrieved information and reduce noise. More sophisticated fusion strategies may also be explored to dynamically balance external knowledge with the intrinsic reasoning of LLMs, enabling more effective information integration.

\bibliography{custom}

\clearpage
\appendix

\section{Appendix}
\label{sec:appendix}
\subsection{Datasets}
\label{sec:dataset}
We utilize two benchmark KGQA datasets, WebQSP~\cite{yih2016value} and CWQ~\cite{talmor2018web}, as proposed in previous studies. Following ROG, we maintain the same training and testing splits. The dataset statistics are provided in Table~\ref{tab:datasets}.
\begin{table}
\centering
\caption{Statistics of datasets.}
\label{tab:datasets}
\begin{tabular}{c|ccc}
\hline
Datasets & \#Train & \#Test & Max \#hop \\
\hline
WebQSP & 2,826 & 1,628 & 2 \\
CWQ & 27,639 & 3,531 & 4 \\
\hline
\end{tabular}
\end{table}
\subsection{Prompt Example}\label{sec:prompt}
\tcbset{
    mybox/.style={
        colback=white,
        colframe=black,
        title=#1,
        fonttitle=\bfseries
    }
}
\begin{figure}[!b]
\begin{tcolorbox}[mybox={Prompts}]
Based on the reasoning paths, please answer the given question. Please keep the answer as simple as possible and return all the possible answers as a list.

\textbf{Reasoning Paths:}

\colorbox{green!20}
{High Priority Paths:}\\
Northern Colorado Bears football $\rightarrow$ education.educational\_institution.sports\_teams $\rightarrow$ University of Northern Colorado

\colorbox{green!20}{Additional Paths:}\\
Northern Colorado Bears football $\rightarrow$ education.educational\_institution.sports\_teams $\rightarrow$ University of Northern Colorado\\
Greeley $\rightarrow$ location.location.containedby $\rightarrow$ United States of America\\
Greeley $\rightarrow$ location.location.containedby $\rightarrow$ Greeley Masonic Temple

\textbf{Question:} 
What educational institution has a football sports team named Northern Colorado Bears is in Greeley, Colorado?
\end{tcolorbox}
\caption{An Example of Our Prompt}
\label{originalprompt}
\end{figure}

\end{document}